\documentclass[letterpaper, 10 pt, conference]{ieeeconf}  % Comment this line out if you need a4paper

\IEEEoverridecommandlockouts                              % This command is only needed if 
\usepackage{url}            % simple URL typesetting
\usepackage{amsmath} 
\usepackage{amssymb}
\usepackage{booktabs}
\usepackage{multirow} 
\usepackage{algorithm}
\usepackage{algorithmic}
\usepackage{xcolor}

\usepackage{threeparttable}
\usepackage{graphicx}

\usepackage{hyperref}  %hyperref still needs to be put at the end!

\usepackage[
backend=biber,
style=numeric-comp,
sorting=none
]{biblatex}
\title{\LARGE \bf
Mitigating Exposure Bias in Score-Based Generation of Molecular Conformations   
}

\author{Sijia Wang$^{1}$, Chen Wang$^{1}$, Zhenhao Zhao$^{1}$, Jiqiang Zhang$^{2}$ and  Weiran Cai$^{1,*}$% <-this % stops a space
\thanks{$^{1}$School of Computer Science and Technology, Soochow University, Suzhou, 215006, China}
\thanks{$^{2}$School of Physics, Ningxia University, Yinchuan, 750021, China}
\thanks{$^{3}$Correspondence should be addressed to: {\tt\small wrcai@suda.edu.cn}} %
}

\begin{document}
\maketitle
\thispagestyle{empty}
\pagestyle{empty}
%%%%%%%%%%%%%%%%%%%%%%%%%%%%%%%%%%%%%%%%%%%%%%%%%%%%%%%%%%%%%%%%%%%%%%%%%%%%%%%%
\begin{abstract}

Molecular conformation generation poses a significant challenge in the field of computational chemistry. Recently, Diffusion Probabilistic Models (DPMs) and Score-Based Generative Models (SGMs) are effectively used due to their capacity for generating accurate conformations far beyond conventional physics-based approaches. However, the discrepancy between training and inference rises a critical problem known as the exposure bias. While this issue has been extensively investigated in DPMs, the existence of exposure bias in SGMs and its effective measurement remain unsolved, which hinders the use of compensation methods for SGMs, including ConfGF and Torsional Diffusion as the representatives. In this work, we first propose a method for measuring exposure bias in SGMs used for molecular conformation generation, which confirms the significant existence of exposure bias in these models and measures its value. We design a new compensation algorithm Input Perturbation (IP), which is adapted from a method originally designed for DPMs only. Experimental results show that by introducing IP, SGM-based molecular conformation models can significantly improve both the accuracy and diversity of the generated conformations. Especially by using the IP-enhanced Torsional Diffusion model, we achieve new state-of-the-art performance on the GEOM-Drugs dataset and are on par on GEOM-QM9. We provide the code publicly at \href{https://github.com/jia-975/geodiff-ip}{https://github.com/jia-975/torsionalDiff-ip}.

\end{abstract}
%%%%%%%%%%%%%%%%%%%%%%%%%%%%%%%%%%%%%%%%%%%%%%%%%%%%%%%%%%%%%%%%%%%%%%%%%%%%%%%%
\section{INTRODUCTION}

Molecular conformation refers to the three-dimensional coordinates of a molecule, which is instrumental in accomplishing many tasks such as molecular property prediction \cite{gilmer2017neural,hu2020Strategies}, docking \cite{roy2015understanding}, and molecule generation \cite{shi2020graphaf, xie2021mars}. Traditional methods such as Molecular Dynamics \cite{de2016role} leverage physical principles to generate relatively accurate conformations. However, these methods regularly suffer from efficiency problems, particularly when dealing with large molecules \cite{hu2021ogb}. Therefore, the generation of stable molecular conformations remains a grand challenge, primarily due to the intricate nature of molecular interactions and the computational demands associated with accurate modeling.

Recent advancement has witnessed deep generative models as a powerful tool for generating molecular conformations. For instance, \cite{mansimov2019molecular} introduces a Conditional Variational Graph Auto-Encoder, which represents the first attempt to use deep generative models for this problem, yet neglects the importance of invariance. To tackle the issue, GraphDG \cite{simm2019generative} and CGCF \cite{xu2021learning} model atomic distances, which are roto-translation invariant. \cite{shi2021learning} estimates the score of atomic distances. DGSM \cite{luo2021predicting} improves the quality by modeling the long-range interactions of non-bonded atoms. GeoDiff, proposed in \cite{xu2022geodiff}, uses equivariant neural networks to achieve the roto-translational equivariance. Furthermore, different from previous models that operate in the Euclidean space, \cite{jing2022torsional} proposes the Torsional Diffusion in the space of torsional angles. 

However, as pointed out in \cite{ddpm-ip}, a systematic discrepancy arises between training and generation phases associated with Diffusion Probabilistic Models (DPMs) -- while the former is conditioned on real samples, the latter is conditioned on generated ones, ultimately leading to the problem known as an exposure bias. \cite{ddpm-ip} first proposes this concept and suggests that a bias reduction approach by perturbing the input in the training phase. The issue of the exposure bias in DPMs has received significant attention, for which several effective methods have been proposed for its alleviation. \cite{li2023alleviating} discovers the existence of the time-shift between training and sampling and proposes a new sampler to mitigate this bias. \cite{ning2023elucidating} points out that the rooting cause of exposure biases is the prediction error at each sampling step and proposes the Epsilon Scaling method. 

In our work, we attempt to examine the exposure bias in molecular conformation generation and seek a solution for its mitigation. However, exposure biases have only been confirmed in DPMs up to date. Whether there is exposure biases in Score-Based Generative Models (SGMs) and how to effectively measure such bias still lacks exploration. This hinders the use of compensation methods in the SGM category, including a full range of important molecular conformation generating models such as Torsional Diff and ConfGF. Here, we propose a method for measuring the exposure bias specifically in SGMs. With the detecting algorithm, we confirm the significant existence of exposure biases in this type of models and are able to make estimations of the values. This result enables the adaptation of a compensation algorithms for DPMs to SGMs. Specifically, our experiments on GEOM-QM9\cite{ramakrishnan2014quantum} and GEOM-Drugs \cite{axelrod2022geom} dataset have confirmed that by adapting the exposure bias compensation method Input Perturbation\cite{ddpm-ip}, originally designed for DPMs, to the representative models ConfGF and Torsional Diffusion in the SGM category can significantly improve both accuracy and diversity of the generation performance. Most notably, by using Torsional Diffusion as a baseline, we achieve a new state-of-the-art performance on the GEOM-Drugs dataset and a performance on par with the SOTA on GEOM-QM9. Overall, our contributions are:
\begin{itemize}
    \item We propose a method for measuring the exposure bias in SGMs, with which we confirm the significant existence of exposure bias in the representative model ConfGF in this category.
    \item We adapt an exposure bias compensation method Input Perturbation for training two SGMs for molecular conformations, namely Torsional Diffusion and ConfGF, and show significant improvement regarding both accuracy and diversity of the generated results. Notably, we achieve a new state-of-the-art performance on the GEOM-Drugs dataset using Torsional Diffusion as a baseline.
\end{itemize}
%%%%%%%%%%%%%%%%%%%%%%%%%%%%%%%%%%%%%%%%%%%%%%%%%%%%%%%%%%%%%%%%%%%%%%%%%%%%%%%%
\section{Related Work} 

In recent years, the employment of SGMs or DPMs in molecular conformation generation has attracted increasing attention. ConfGF introduces an innovative model that employs SGMs for molecular conformation \cite{shi2021learning}. ConfGF predicts scores for molecular distances and then computes conformation scores via the chain rule. The estimated scores allow generating stable conformations via the Langevin dynamics. \cite{luo2021predicting} points out that we should not only model the forces between bonded atoms, but also consider those between non-bonded atoms. They propose the DGSM model, which further improves the quality of generated conformations. On the other hand, \cite{xu2022geodiff} designs an equivariant neural network \cite{satorras2021n} to address the roto-translation equivariance. Each atom is viewed as a particle in GeoDiff, which learns to reverse the diffusion process as a Markov chain. Furthermore, \cite{jing2022torsional} proposes the Torsional Diffusion by using a torsion angle space. This model divides conformations into internal coordinates, computed using standard cheminformatic methods, and torsion angles, obtained through a supermanifold-defined SGMs. Torsional Diffusion significantly reduces modeling complexity and enables the generation of high-quality conformations in shorter time. 

However, DPMs such as GeoDiff have a common problem: they use real samples during training as inputs, but use predicted samples during generation, leading to an exposure bias \cite{ddpm-ip, li2023alleviating, ning2023elucidating}. The issue has received significant attention in DPMs, and several effective methods have been proposed to alleviate this bias. \cite{ddpm-ip} first verifies that this bias follows a Gaussian distribution. They introduce a method named Input Perturbation (IP) by perturbing the input for reducing the bias. Time-Shift \cite{li2023alleviating}  locates a more compatible time point which can reduce the bias and proposes the Time-Shift Sampler. Epsilon Scaling \cite{ning2023elucidating} shows that the network output derived from real samples is always less than that obtained from predicted ones. However, the exposure bias in SGMs is still not tackled for molecular conformation generation. A reliable method for measuring exposure bias in SGMs is in need, which  should confirm the existence of exposure biases in the models in this category and estimate their values. 

%%%%%%%%%%%%%%%%%%%%%%%%%%%%%%%%%%%%%%%%%%%%%%%%%%%%%%%%%%%%%%%%%%%%%%%%%%%%%%%%
\section{Preliminaries}
\label{sec:Preliminaries}
%%%%%%%%%%%%%%%%%%%%%%%%%%%%%%%%%%%
\subsection{Problem Definition}

\noindent\textbf{Notations.} In this paper, we represent a molecular graph as \(G = \langle V, E \rangle\), where \(V = \{v_1, v_2, \dots, v_{|V|}\}\) is the set of vertices representing atoms, and \(E = \{ e_{ij} \mid (i, j) \subseteq V \times V \}\) is the set of edges representing chemical bonds in the molecule. Each node \(v_i\) represents atomic properties, such as nuclear charge and atomic coordinate \(c_i \in \mathbb{R}^3 \), and each edge \(e_{ij}\) represents the properties of the chemical bond between atoms \(v_i\) and \(v_j\), such as the type of the bond. 

Since the directly connected bond relationships are insufficient to fully characterize the conformation, we introduce virtual edges between 2-hop and 3-hop neighbors, as suggested in ConfGF \cite{shi2021learning} and GeoDiff \cite{xu2022geodiff}, to reduce the degree of freedom. Hereafter, unless otherwise specified, we assume all molecular graphs are extended.

\noindent\textbf{Problem Definition.} Given a graph $G = \langle V, E \rangle$, our objective is to establish a
model that generates corresponding 3D molecular conformations $C \in \mathbb{R}^{|V| \times 3}$.
%%%%%%%%%%%%%%%%%%%%%%%%%%%%%%%%%%%
\subsection{Score-based Generative Models} 
Score-based generative models (SGMs) \cite{song2019generative, song2020score, song2020improved} are a class of generative models that estimate the gradient of the data distribution via denoising score matching, and use annealed Langevin dynamics  \cite{welling2011bayesian} for sampling. The objective minimizes $\frac{1}{2} E_{p_{data}(C_0)} [\| s_{\theta}(C_0) - \nabla_{C_0} \log p_{data}(C_0) \|^2_2]$, which can be equivalently represented in the following:
\begin{equation}
 E_{p_{data}(C_0)}[  \text{tr}(\nabla_{C_0}s_\theta(C_0)) + \frac{1}{2}\| s_\theta(C_0) \|^2_2],
\end{equation}
where $C_0$ denotes the original data and $\text{tr}(\nabla_{C_0}s_\theta(C_0))$ is the Jacobian of $s_\theta(C_0)$, which is not scalable. However, the difficulty can potentially be solved by Denoising Score Matching \cite{vincent2011connection}, which perturbs the data $C_0$ and estimates the score of the perturbed data distribution instead. To be specific, for time $t$ and a noise level $\sigma_t$, the noised sample $C_t$ can be expressed by
\begin{equation}
\label{equ: add noise}
C_t = C_0 + \sigma_t\epsilon,
\end{equation}
where $\epsilon \sim N(0,1)$. The objective is then expressed as
\begin{equation}
\label{equ:loss}
\frac{1}{2} \mathbb{E}_{p(C_0)} \mathbb{E}_{q_{\sigma_t}(C_t \mid C_0)} \left[ \| s_\theta(C_t, \sigma_t) - \nabla_{C_t} \log q_{\sigma_t}(C_t \mid C_0) \|_2^2 \right].
\end{equation}
When $\sigma_t \approx 0$, we consider $s_\theta(C_t, \sigma_t) \approx \nabla_{C_0}\log p(C_0)$.   
After the  score networks $ s_\theta(C_t, \sigma_t)$ are trained, we can sample using annealed Langevin dynamics by continuously reducing the noise level. The sampling algorithm is shown in Algorithm \ref{alg:anneal sampling} %\textcolor{blue}{(This algorithm would be better put after the training algorithm?)}.

\begin{algorithm}[H]
    %\vspace{-4pt}
	\caption{Score model for conformation sampling}
	\label{alg:anneal sampling}
    \renewcommand\algorithmiccomment[1]{\hfill $\triangleright$ {#1}}	
	\begin{algorithmic}[1]
	    \STATE \textbf{Input: }{molecular graph $G$, noise levels $\{\sigma_t\}_{t=1}^L$, the smallest step size $a$, and the number of sampling steps per noise level $T$.}
	    %\COMMENT{$\epsilon$ is the smallest step size, and $T$ is the number of sampling steps per noise level.}
	    \STATE{Initialize conformation $ {\hat{C}_T}$ from a prior distribution}
	    \FOR{$t \gets L$ to $1$}
	        \STATE{$\alpha_t \gets a \cdot \sigma_t^2/\sigma_L^2$} 
            \FOR{$i \gets 1$ to $T$}
                \STATE{Draw ${z}_i \sim \mathcal{N}(0, {I})$}
                \STATE{${\hat{C}_i} \gets {\hat{C}_{i-1}} + \alpha_t {s}_{\theta}({\hat{C}_{i-1}}, \sigma_t) + \sqrt{2 \alpha_t}{z}_i$}
            \ENDFOR
            \STATE{${\hat{C}_0} \gets {\hat{C}_T}$}
        \ENDFOR
        \STATE \textbf{return} ${\hat{C}_0}$
	\end{algorithmic}
    %\vspace{-5pt}
\end{algorithm}
%%%%%%%%%%%%%%%%%%%%%%%%%%%%%%%%%%%%%%%%%%%%%%%%%%%%%%%%%%%%%%%%%%%%%%%%%%%%%%%%
\section{Exposure Bias in Conformation Generation}
%%%%%%%%%%%%%%%%%%%%%%%%%%%%%%%%%%%
\subsection{Estimating Exposure Bias in SGMs}

The concept of exposure bias is originally coined for DPMs, which addresses the discrepancy between training and inference phases \cite{ddpm-ip}. Whether there is a similar exposure bias in SGMs and how to effectively measure it remains a key issue, which can significantly affect the conformation quality generated by this type of models, including ConfGF and Torsional Diffusion as representatives. Indeed, during the training process of SGMs, the input \({C_t}\) to the network can be regarded as a ground truth sample (Eq. \ref{equ: add noise}), while during the inference, the input is the predicted sample $\hat{C_{t}}$ along the Markov chain. This discrepancy between the ground truth and predicted samples accumulates in the sampling chain in the generation process, which should finally lead to a similar exposure bias as in DPMs \cite{ddpm-ip, li2023alleviating, ning2023elucidating}. 

In order to effectively measure the accumulation of such a bias in SGMs, we use the true sample $C_0$ to compare with the corresponding generated final result $\hat{C}_0$. Specifically, at a noise level $\sigma_t$, we first obtain $C_t$ with a noise adding function equation $f_N$, which can be expressed by $C_t = f_N(C_0)$ with a noise level $\sigma_t$ (which in the case of Confgf can be simply written as Eq. \ref{equ: add noise}). Then we use $C_t$ as a starting point to generate $\hat{C}_0$ but with a zero-noise sampling function equation $f_S$ that does not include the random term: ${\hat{C}_t} = f_S({\hat{C}_{t-1}}, \alpha_t {s}_{\theta}({\hat{C}_{t-1}}, \sigma_t))$. Based on these two values, the estimate of the exposure error is computed as $|e_t| =||\hat{C}_0 - C_0||$. The algorithm is summarized in Algorithm \ref{alg: error}. 

\begin{algorithm}[h]
   \caption{Exposure bias in SGMs for conformations}
   \label{alg: error}
   \renewcommand\algorithmiccomment[1]{\hfill $\triangleright$ {#1}}	
   \begin{algorithmic}[1]
        \STATE \textbf{Input:} molecular graph $G$, noise levels $\{\sigma_t\}_{t=1}^L$, function $f_N$ used to obtain $C_t$,
        zero-noise sampling function equation $f_S$, and pre-trained score model $s_\theta$.
        \STATE Initialize $e_t = 0$, $n_t = 0$ ($\forall t \in \{1, ..., L \}$)
        \REPEAT
        \STATE $C_0 \sim q(C_0)$, $t\sim \mathbb{U}(\{1,...,L\})$, ${\epsilon} \sim {\cal N} ({0}, {I})$
        % \FOR{}
        \STATE $C_t \gets f_N(C_0, \sigma_t, \epsilon)$
        \COMMENT{obtain ground truth $C_t$}
        \FOR{$t \gets t-1$ to $1$}
            \STATE{${\hat{C}}_{t-1} \gets f_S({\hat{C}}_{t}, {s}_{\theta}({\hat{C}}_{t}, \sigma_t)$}
        \ENDFOR
        \STATE $|e_t| = |e_t| + ||{C}_{0} - \hat{{C}}_{0}||_1/N$  \COMMENT{$N$ is the number of atoms in ${C}_{0}$}
        \STATE $n_t = n_t + 1$
        \UNTIL {$N$ iterations}
        \STATE if $n_t \neq 0$, then  $|\bar{e_t}| = \frac{|e_t|}{n_t}$ ($\forall t \in \{1, ..., L \}$)
    \end{algorithmic}
\end{algorithm}

\begin{figure}[ht]
\vskip -0.15in
\begin{center}
\centerline{\includegraphics[scale=0.35, trim=0pt 0pt 0pt 0pt, clip]{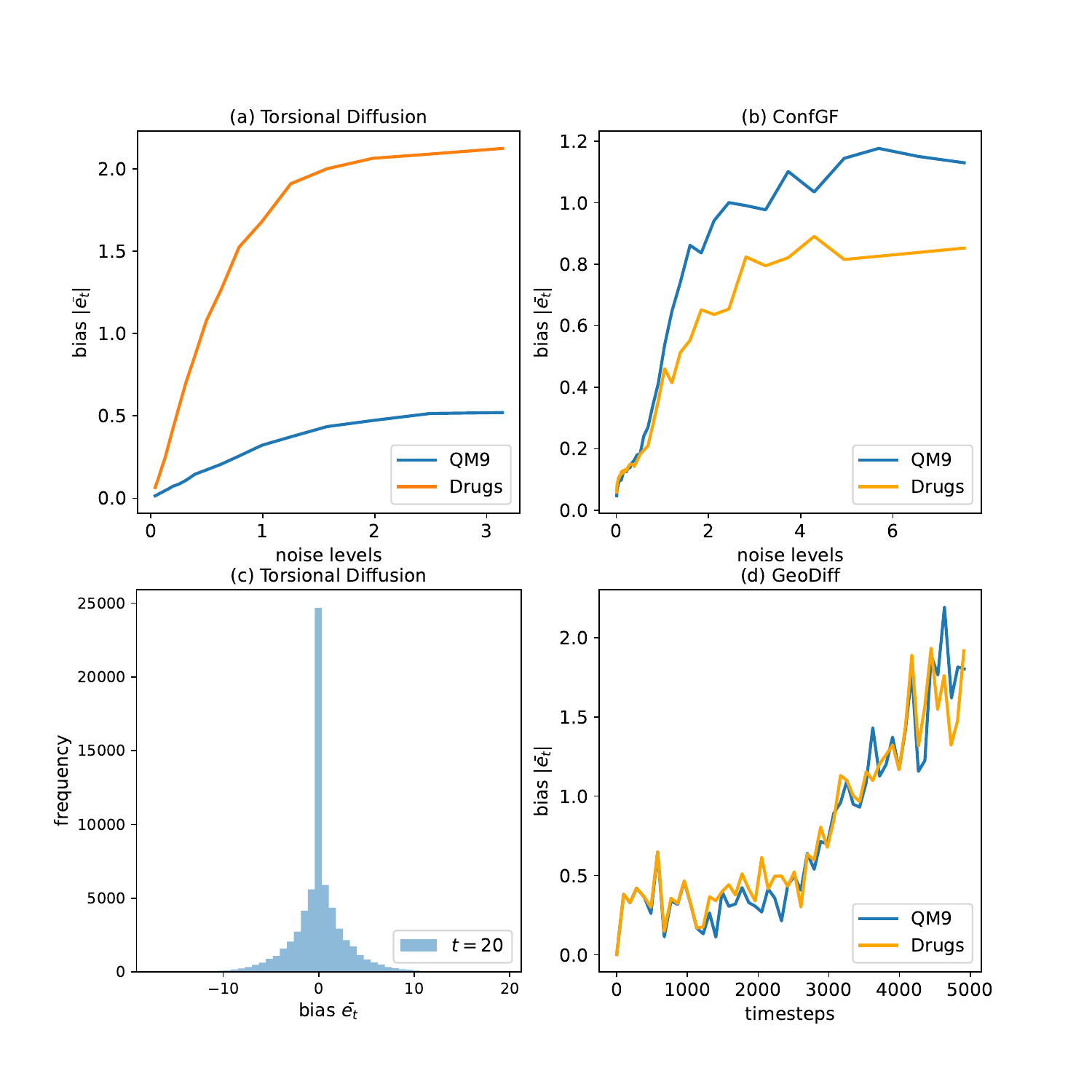} } 
\caption{Exposure bias estimated with pre-trained (a) ConfGF, (b) Torsional Diffusion and (d) GeoDiff over 5000 samples on both QM9 and Drugs datasets. (c) Empirical distribution of $e_t$ with Torsional Diffusion.}
\label{pic: bias}
\end{center}
\vskip -0.2in
\end{figure}

We use the representative molecular conformation generation models ConfGF \cite{song2019generative} and Torsional diffusion \cite{jing2022torsional} in the SGM category for the bias detection experiment. For the QM9 and Drugs datasets, we used 5000 samples for generation, where each sample corresponds to 5 conformations. The results are presented in Fig .\ref{pic: bias}. We observe that the exposure bias gradually decreases as the noise level decreases. Specifically, for the QM9 dataset, we observe that the detected average bias is 0.39, while in the Drugs dataset, the average bias is 0.29, which provides important estimates for the subsequent compensation. Therefore, the proposed measurement method has confirmed the existence of exposure bias in the SGMs, which has laid the foundation for the future bias compensation. We have also estimated the exposure error of GeoDiff in the DDPM category, following the method introduced in \cite{ddpm-ip}, where the bias curve exhibits a similar trend with those for the SGM models but with a concavity.

%%%%%%%%%%%%%%%%%%%%%%%%%%%%%%%%%%%
\subsection{Input Perturbation}

The estimated bias confirms the existence of an exposure bias in SGMs, which requires a compensation method for its alleviation. As demonstrated in Fig. \ref{pic: bias}c, as well as in \cite{ddpm-ip}, the bias between the predicted and the true distributions follows a Gaussian-like distribution (though not in a strict sense). Therefore, to mitigate the inconsistency between training and generation phases, we perturb the input with a Gaussian noise in the training of the score function to enhance its robustness \cite{ddpm-ip}. Specifically, we introduce perturbations to \(C_0\) in Eq. \ref{equ: add noise}
\begin{equation}
\label{equ: calculate Ct with ip}
\tilde{C_t} = C_0 + \sigma_t(\epsilon + \lambda_t\xi)
\end{equation}
where $\lambda_t$ is the weight of perturbation. The training objective remains $q_{\sigma_t}(C_t | C_0)$, but the input is set to $\tilde{C_t}$ to take into account different conformations  other than the ground truth, which is detailed in Algorithm \ref{alg: ip-training}. The generation process remains unchanged, as shown in Algorithm \ref{alg:anneal sampling}. This asymmetric procedure thus allows the trajectories to stick closer to the ground truth despite potential biases during generation. Here, we also follow the routine as in \cite{ddpm-ip} and use a uniform noise level by setting $\lambda_t = \lambda$ for perturbation, independently of t, to avoid the complexity of training it twice. This is notable that since we only perturb the input, the rotational and translational equivariance is not affected. $\lambda$ is empirically set using a grid search on both QM9 and Drugs on a small range of values covering the last half of the sampling trajectory (as in Fig.\ref{pic: bias}), which has usually the largest impact on the performance.
\begin{algorithm}[h]
   \caption{Training of SGMs with Input Perturbation}
   \label{alg: ip-training}
   \renewcommand\algorithmiccomment[1]{\hfill $\triangleright$ {#1}}	
   \begin{algorithmic}[1]
        \REPEAT
            \STATE $C_0 \sim q(C_0)$, $t\sim \mathbb{U}(\{1,...,L\})$, ${\epsilon} \sim {\cal N} ({0}, {I})$
            \STATE $C_t = C_0 + \sigma_t\epsilon$ 
            \STATE $\tilde{C_t} = C_0 + \sigma_t(\epsilon + \lambda_t\xi)$ 
            % \STATE{${d}_{t} = g_G({C}_{t})$}
            \STATE $\text{gradient descent step on} \: \nabla_{ {\theta}} || {s_\theta(\tilde{C_t}, \sigma_t)} + \frac{C_t - C_0}{\sigma_t ^2} ||^2$  
        \UNTIL {converged}
    \end{algorithmic}
\end{algorithm}

As shown in Table \ref{tab: exposure-bias estimate}, the model trained with the Input Perturbation exhibits a smaller exposure bias, which validates the effectiveness of the compensation method. We can observe that the exposure bias $|\bar{e_t}|$ increases with $\sigma_t$, which is attributed to that more denoising steps may be needed for a larger $\sigma$, leading to an accumulated bias.
\begin{table}[ht]
    \centering
    \caption{Empirical estimate of exposure bias on GEOM-Drugs.}
    \label{tab: exposure-bias estimate}
    \begin{tabular}{lcccccc}
        \toprule 
        $\sigma_t$ & 0.02 & 0.05 & 0.09 & 0.19 & 0.39 & 0.79  \\
        \midrule
        ConfGF     & 0.072 & 0.086 & 0.098 & 0.110 & 0.136 & 0.301 \\
        ConfGF-IP  & 0.065 & 0.077 & 0.088 & 0.097 & 0.126 & 0.285 \\
        \bottomrule
    \end{tabular}
\end{table}
%%%%%%%%%%%%%%%%%%%%%%%%%%%%%%%%%%%%%%%%%%%%%%%%%%%%%%%%%%%%%%%%%%%%%%%%%%%%%%%%
\section{Experiments}

\subsection{Experiments Setup}
\noindent\textbf{Data} Following CGCF \cite{xu2021learning}, we use the GEOM-QM9\cite{ramakrishnan2014quantum} and GEOM-Drugs\cite{axelrod2022geom} datasets for our experiments. The GEOM-QM9 dataset, derived from the QM9 dataset, comprises 133,258 molecules, each containing an average of 9 heavy atoms per molecule. The GEOM-Drugs dataset has 304,466 molecules, each containing an average of 24.9 heavy atoms per molecule. We adopt the data partitioning method introduced in ConfGF \cite{shi2021learning} and GeoDiff \cite{xu2022geodiff}, where 40,000 molecules are sampled for the training set. For each molecule, we select 5 conformations with low energy, resulting in a total of 200,000 conformations in the training set. The test set consists of 200 molecules. As the partitioning method for training set and test set of Tor. Diff. differs from that of typical deep learning models, we have therefore retrained the model. 

\noindent\textbf{Baseline} We compare our method with state-of-the-art deep learning models. Specifically, they include VAE based CVGAE \cite{mansimov2019molecular}, GraphDG \cite{simm2019generative} and ConfVAE \cite{xu2021confVAE}, Continuous Flow based CGCF \cite{xu2021learning}, GeoMol \cite{ganea2021geomol}, DPM-based GeoDiff \cite{xu2022geodiff}, SGM-based ConfGF\cite{shi2021learning},  DGSM\cite{luo2021predicting}, and Torsional Diffusion \cite{jing2022torsional}. We denote our input perturbation with the $\{\cdot\}$-IP to indicate their relationship to the baseline methods. For example, the input perturbation of ConfGF is refered to as ConfGF-IP. 
%%%%%%%%%%%%%%%%%%%%%%%%%%%%%%%%%%%
\subsection{Conformation Comparison}
\label{conformation generation}

\noindent\textbf{Setup} We compare the quality of generated conformations. In our experiments, we generate twice the number of conformations as in the test set. To measure the difference between true and generated conformations, we employ the Root Mean Square Deviation (RMSD) for the heavy atoms: 
\begin{equation}
    \label{RMSD}
    \text{RMSD}(C, \hat{C}) = \min_{\Phi} \left( \sqrt{\frac{1}{n} \sum_{i=1}^{n} \left\| \Phi(C_i) - \hat{C}_i \right\|^2} \right)
\end{equation}
where $n$ is the number of heavy atoms, \(\phi\) is an alignment function using Kabsch algorithm\cite{Kabsch:a12999}, $C$ and $\hat{C}$ are the sets of real and generated samples respectively. Following CGCF \cite{xu2021learning}, we utilize the Coverage (COV) and Matching (MAT) to measure diversity and accuracy respectively, which can be defined as:
\begin{equation}
    \label{COV}
    COV(S_g, S_r) = \frac{\left| \{ C \in S_r \, | \, RMSD(C, \hat{C}) < \delta, \hat{C} \in S_g \} \right|}{|S_r|} 
\end{equation}

\begin{equation}
     MAT(S_g, S_r) = \frac{1}{|S_r|} \sum_{C \in S_r} \min_{\hat{C} \in S_g} RMSD(C, \hat{C})
    \label{MAT}
\end{equation}
where \(S_r\) is the set of conformations derived from real data,  and \(S_g\) is the set of generated conformations. A higher COV indicates greater diversity, while a smaller MAT signifies more accurate conformations. The threshold $\delta$ was set to 0.5$\mathrm{\mathring{A}}$ for GEOM-QM9 and 1.25 $\mathrm{\mathring{A}}$ for GEOM-Drugs.

\begin{table}[ht]
    \centering
    \caption{COV and MAT scores on the GEOM-QM9}
    \label{tab:qm9 results}
    \begin{tabular}{lcccc}
        \toprule
        \multirow{2}{*}{Method}&\multicolumn{2}{c}{COV(\%)($\uparrow$)}&\multicolumn{2}{c}{MAT($\mathrm{\mathring{A}}$)($\downarrow$)}\\
        \cmidrule(lr){2-3}   
        \cmidrule(lr){4-5}   
         & Mean &Median & Mean &Median\\    
        \midrule
        RDKit & 83.26 & 90.78 & 0.3447 & 0.2935 \\
        CVGAE & 0.09 & 0.00 & 1.6713 & 1.6088 \\
        GraphDG & 73.33 & 84.21 & 0.4245 & 0.3973 \\
        CGCF & 78.05 & 82.48 & 0.4219 & 0.3900 \\
        ConfVAE & 77.84 & 88.20 & 0.4154 & 0.3739 \\
        GeoMol & 71.26 & 72.00 & 0.3731 & 0.3731 \\
        ConfGF & 88.49 & 94.13 & 0.2673 & 0.2685 \\
        GeoDiff & 90.07 & 93.39 & 0.2090 & 0.1988 \\
        DGSM & \textbf{91.49} & 95.92 & 0.2139 & 0.2137 \\
        Tor. Diff. & 83.07 & \textbf{100.00} & 0.2379 & 0.2085\\
        \midrule
        Tor. Diff.-IP & 84.49 & \textbf{100.00} & \textbf{0.1924} & \textbf{0.1530} \\
        \bottomrule
    \end{tabular}
    \begin{tablenotes}
        \small
        \item * The results of CVGAE, GraphDG, CGCF, ConfVAE and GeoMol are borrowed from GeoDiff \cite{xu2022geodiff}. The results of RDKit, DGSM are borrowed from \cite{luo2021predicting}. The rest are obtained from our own experiments and Tor. Diff. is retrained.
    \end{tablenotes}
\end{table}
\begin{table}[ht]
    \centering
    \caption{COV and MAT scores for GEOM-Drugs}
    \label{tab:drugs results}
    \begin{tabular}{lcccc}
        \toprule
        \multirow{2}{*}{Method}&\multicolumn{2}{c}{COV(\%)($\uparrow$)}&\multicolumn{2}{c}{MAT($\mathrm{\mathring{A}}$)($\downarrow$)}\\
        \cmidrule(lr){2-3}   
        \cmidrule(lr){4-5}   
         & Mean &Median & Mean &Median\\    
        \midrule
        RDKit & 60.91 & 65.70 & 1.2026 & 1.1252 \\
        CVGAE & 0.00 & 0.00 & 3.0702 & 2.9937 \\
        GraphDG & 8.27 & 0.00 & 1.9722 & 1.9845 \\
        CGCF & 53.96 & 57.06 & 1.2487 & 1.2247 \\
        ConfVAE & 55.20 & 59.43 & 1.2380 & 1.1417 \\
        GeoMol & 67.16 & 71.71 & 1.0875 & 1.0586 \\
        ConfGF & 62.15 & 70.93 & 1.1629 & 1.1596 \\
        GeoDiff & 88.36 & 96.09 & 0.8704 & 0.8628 \\
        DGSM & 78.73 & 94.39 & 1.0154 & 0.9980 \\
        Tor. Diff. & 91.63 & \textbf{100.00} & 0.6862 & 0.6641\\
        \midrule
        Tor. Diff.-IP &\textbf{93.26} & \textbf{100.00} & \textbf{0.6731} & \textbf{0.6328} \\
        \bottomrule
    \end{tabular}
    
\end{table}
%%%%%%%%%%%%%%%%%%%%%%%%%%%%%%%%%%%

\noindent\textbf{Results} 
As shown in Tab.\ref{tab:qm9 results} and \ref{tab:drugs results}, using Input Perturbation method in Torsional Diffusion can significantly improve both accuracy and diversity of the generated conformations. For GEOM-QM9, we improve MAT-mean and MAT-median by 19.13\% and 26.62\%, respectively. Notably, for GEOM-Drugs we outperform all previous methods, achieving a new state-of-the-art performance. Experiments show that the Input Perturbation method can enhance the robustness of the model, allowing the sampling to proceed in the correct direction even with a certain deviation and hence reducing the exposure bias.
%%%%%%%%%%%%%%%%%%%%%%%%%%%%%%%%%%%
\subsection{Ablation Study}
To demonstrate the effectiveness of the Input Perturbation method for SGMs, we conduct experiments on both distance-based ConfGF in the Euclidean space and torsion-angle based Torsional Diffusion. In addition, we conduct experiments on the conformation-based GeoDiff, a representative of DPMs, as a reference to compare the performance enhancement on the two types of models. 

\noindent\textbf{Conformation Generation.} The quality of conformation generation using IP is compared for three baseline models as shown in Tab. \ref{tab:ablation results}. In both Euclidean space and torsional angle space, the Input Perturbation method consistently improve the accuracy and diversity of the conformation. Futhermore, the experiments have confirmed that compared with the IP-based DPMs, we have achieved a considerable enhancement in the performance enhancement with IP-based SGMs. 
\begin{table}[ht]
    \centering
    \caption{COV and MAT scores for GEOM-QM9 and GEOM-Drugs}
    \label{tab:ablation results}
    \begin{tabular}{llccccc}
        \toprule
        \multirow{2}{*}{Data} & \multirow{2}{*}{Method} & \multicolumn{2}{c}{COV(\%)($\uparrow$)} & \multicolumn{2}{c}{MAT($\mathrm{\mathring{A}}$)($\downarrow$)} \\
        \cmidrule(lr){3-4}
        \cmidrule(lr){5-6}
         & & Mean & Median & Mean & Median\\    
        \midrule
        \multirow{6}{*}{QM9} 
        
        & GeoDiff & 90.07 & 93.39 & 0.2090 & 0.1988 \\
        & GeoDiff-IP & \textbf{90.13} & \textbf{95.20} & \textbf{0.2086} & \textbf{0.1970} \\
       
        \cmidrule(lr){2-6}
         & ConfGF & 88.49 & 94.13 & 0.2673 & 0.2685 \\
        & ConfGF-IP & \textbf{91.28} & \textbf{95.80} & \textbf{0.2655} & \textbf{0.2604} \\
        \cmidrule(lr){2-6}
         &Tor. Diff. & 83.07 & \textbf{100.00} & 0.2379 & 0.2085\\
        &Tor. Diff.-IP & \textbf{84.49} & \textbf{100.00} & \textbf{0.1924} & \textbf{0.1530} \\
        \midrule
        \multirow{6}{*}{Drugs} 
        & GeoDiff & 88.36 & 96.09 & 0.8704 & 0.8628 \\
        & GeoDiff-IP & \textbf{89.12} & \textbf{96.33} & \textbf{0.8641} & \textbf{0.8395}  \\
        \cmidrule(lr){2-6}
        & ConfGF & 62.15 & 70.93 & 1.1629 & 1.1596 \\
        & ConfGF-IP & \textbf{64.05} & \textbf{74.83} & \textbf{1.1522} & \textbf{1.1236} \\
        \cmidrule(lr){2-6}
        &Tor. Diff. & 91.63 & \textbf{100.00} & 0.6862 & 0.6641\\
        &Tor. Diff.-IP &\textbf{93.26} & \textbf{100.00} & \textbf{0.6731} & \textbf{0.6328} \\
        \bottomrule
    \end{tabular}
\end{table} 

\textbf{Property Prediction.} The task of property prediction involves predicting molecular ensemble properties \cite{axelrod2022geom} over a set of generated conformations. We sample 30 new molecules from the QM9 dataset and generate 50 conformations for each molecule. Quantum chemistry calculations are performed to compute the energy \(E\), HOMO-LUMO gap \(\Delta \varepsilon\) for each conformation. For ConfGF-IP and GeoDiff-IP, we follow ConfGF \cite{shi2021learning} and GeoDiff \cite{xu2022geodiff} to calculate using Psi4 \cite{psi4}. For Torsional Diffusion-IP, we follow Torsional Diffusion \cite{jing2022torsional} and use GFN2-xTB\cite{bannwarth2019gfn2}. Subsequently, we compare the average energy \(\overline{E}\), lowest energy $E_{\text{min}}$, average gap $\overline{\Delta \varepsilon}$, minimum gap $\Delta \varepsilon_{\text{min}}$, and maximum gap $\Delta \varepsilon_{\text{max}}$ with the corresponding ground truth values. The mean absolute error (MAE) is used to measure the accuracy of ensemble property prediction. As shown in Table \ref{tab:redicted ensemble properties}, our model significantly outperforms the baselines regarding all five properties.
%\(\overline{E}\), $E_{\text{min}}$, $\overline{\Delta \varepsilon}$ and $\Delta \varepsilon_{\text{min}}$. 

\begin{table}[ht]
    \centering
    \caption{MAE of predicted ensemble properties in eV.}
    \label{tab:redicted ensemble properties}
    \begin{tabular}{lccccc}
        \toprule 
        Method & $\overline{E}(\downarrow) $ & $E_{\text{min}}(\downarrow)$ & $\overline{\Delta \varepsilon}(\downarrow)$ & $\Delta \varepsilon_{\text{min}}(\downarrow)$ & $\Delta \varepsilon_{\text{max}}(\downarrow)$ \\
        \midrule
        GeoDiff & 0.2597 & 0.1551 & 0.3091 & 0.7033 & 0.1909 \\
        GeoDiff-IP &  \textbf{0.2049} & \textbf{0.1029} & \textbf{0.2408} & \textbf{0.4610} & 
        \textbf{0.1311} \\ 

        \midrule
        ConfGF & 2.7886 & 0.1765 & \textbf{0.4688} & 2.1843 & 0.1433 \\
        ConfGF-IP &  \textbf{1.7683} & \textbf{0.1738} & 0.4770 & \textbf{2.0928} & \textbf{0.1289} \\

        \midrule
        Tor.Diff. & 1.5077 & 3.0301 & 2.3974 & 11.9655 & 6.5355 \\

        Tor.Diff.-IP & \textbf{0.6921} & \textbf{0.4009} & \textbf{1.8476} & \textbf{1.7586} & \textbf{2.6722} \\
        \bottomrule
    \end{tabular}
\end{table} 
%%%%%%%%%%%%%%%%%%%%%%%%%%%%%%%%%%%%%%%%%%%%%%%%%%%%%%%%%%%%%%%%%%%%%%%%%%%%%%%%
\section{Conclusion}

In this paper, we propose a method that estimates the exposure bias in SGMs used for molecular conformation generation, which is caused by the discrepancy existing between the inputs of the training and generating phases in the diffusion model, which is originally exhibited in DPMs. Based on the confirmation of the exposure bias, we adapt the bias compensation algorithms Input Perturbation for DPMs to SGMs, which significantly improve both accuracy and diversity of the generated molecular conformations. This demonstrates the necessity of alleviating such accumulated bias in the conformation generation and our method has extended both the concept of this type of systematic bias and an effective approach for their compensation. 
%%%%%%%%%%%%%%%%%%%%%%%%%%%%%%%%%%%%%%%%%%%%%%%%%%%%%%%%%%%%%%%%%%%%%%%%%%%%%%%%
%%%%%%%%%%%%%%%%%%%%%%%%%%%%%%%%%%%

%%%%%%%%%%%%%%%%%%%%%%%%%%%%%%%%%%%%%%%%%%%%%%%%%%%%%%%%%%%%%%%%%%%%%%%%%%%%%%%%
\printbibliography
\addtolength{\textheight}{-12cm}   % This command serves to balance the column lengths

%%%%%%%%%%%%%%%%%%%%%%%%%%%%%%%%%%%%%%%%%%%%%%%%%%%%%%%%%%%%%%%%%%%%%%%%%%%%%%%%

\end{document}